\documentclass{article}
\usepackage{ijcai21}

\usepackage{times}

\usepackage{soul}
\usepackage{url}
\usepackage[hidelinks]{hyperref}
\usepackage[utf8]{inputenc}
\usepackage[small]{caption}
\usepackage{graphicx}
\usepackage{amsmath,amssymb,amsfonts}
\usepackage{algpseudocode}
\usepackage{algorithm}
\usepackage{booktabs}
\usepackage{subcaption}
\title{Automated Label Generation for Time Series Classification with Representation Learning: Reduction of Label Cost for Training}

\author{
Soma Bandyopadhyay\footnote{Contact Author}\and
Anish Datta\And
Arpan Pal\\
\affiliations
TCS Research, TATA Consultancy Services ,Kolkata, India \\
\emails
\{soma.bandyopadhyay, anish.datta, arpan.pal\}@tcs.com
}

\begin{document}

\maketitle

\begin{abstract}
Time-series generated by end-users, edge devices, and different wearables are mostly unlabelled. We propose a method to auto-generate labels of un-labelled time-series, exploiting very few representative labelled time-series. Our method is based on representation learning using Auto Encoded Compact Sequence (AECS) with a choice of best distance measure. It performs self-correction in iterations, by learning latent structure, as well as synthetically boosting representative time-series using Variational-Auto-Encoder (VAE) to improve the quality of labels. We have experimented with UCR and UCI archives, public real-world univariate, multivariate time-series taken from different application domains. Experimental results demonstrate that the proposed method is very close to the performance achieved by fully supervised classification. The proposed method not only produces close to benchmark results but outperforms the benchmark performance in some cases. 
\end{abstract}

\section{Introduction}

Auto-generation of labels is an open area of research. We know self annotation is a tedious task, and annotations done by experts are costly. The main aim of this research area is to reduce the labelling cost of training data, and, at the same time, minimize domain expert's involvement.  

In this work, we propose a novel mechanism to auto-generate labels. We use a small amount (2\% to 15\%) of representative labelled time-series ($\pounds$). Validation of this small amount of data by subject experts is always less time-consuming and less expensive. Our prime aim is to reduce labelling costs. Proposed Multi-stage Label Generation uses learned Auto Encoded Compact Sequence (AECS) with choice of best distance measure \cite{bandyopadhyay2021}. It performs self-correction in iterations, by using a Variational Auto-encoder (VAE) based generative model, to improve the purity of labels as one of the key functionalities. We perform extensive analysis using univariate and multi-variate 1-D sensor time-series from diverse applications like Healthcare (ECG), Human Activity Recognition (Accelerometer, Camera), Smart-city (Electric Meter), etc. 

Our prime contributions are :-

(1) Neighborhood mapping using the choice of best distance measure: Associating best match w.r.t $\pounds$ and unlabelled time-series ($X_u$; $u$ denotes $unlabelled$), by using learned Auto Encoded Compact Sequence (AECS) of $X_u$. Learned latent representation AECS, has a length much less than the original time-series. Agglomerative hierarchical clustering (HC) is applied on AECS by exploiting appropriate distance measures like Chebyshev (CH), Manhattan (MA), and Mahalanobis (ML). Additionally, HC-AECS \cite{bandyopadhyay2021} uses Modified Hubert statistic ($\mathcal{T}$) \cite{hubert1985}, as an internal clustering measure. We rank clustering formed by different distance measures, based on $\mathcal{T}$, and select the best clustering having highest $\mathcal{T}$ and corresponding distance measure as best distance measure. Association of clusters and corresponding labels of $\pounds$ is performed by applying the best distance measure and using cluster centroids, which are computed using HC on learned AECS.

(2) Self-Correction: Improving quality of generated labels $L_{g_i}$ in iterations, by considering amount of deviations of generated labels between two consecutive iterations $L_{g_{i}}$ and $L_{g_{i+1}}$, boosted by synthetically generated equal amount of $\pounds$. We use Variational Autoencoder (VAE), a generative model, to learn a latent representation of \{$\pounds$\}, and to generate data \{$\pounds_g$\}. Hence, in successive iterations \{$\pounds$\} is expanded, and is getting exploited to learn incrementally and aiding the removal of noise in generated labels. Generation of \{$\pounds$\}, is maintained by reward, in terms of judging significant variations in the generated labels between two successive iterations. This reward is triggered by the label discriminator function, till saturation is reached with a very small tolerance.

(3) Extensive analysis using real-world time-series: We performed the analysis with univariate, multivariate, and variable-length time-series taken from UCR Time Series Classification Archive \cite{dau2019ucr} and UCI Machine Learning Repository \cite{asuncion2007uci}. We used a small amount(5\% to 15\%) of representative labelled time-series \{$\pounds$\}, to generate labels for unlabelled time-series. Performance of benchmark classifiers (like BOSS, Random Forest, MLP, etc.) trained using actual labels differs, on average, only 1.08\% from the classifiers trained with generated labels using our proposed model. A notable point we observe, in some time-series, the performance of the classifier trained with generated labels by our proposed method even exceeds the performance of classifier trained with actual training labels, thereby giving a notion of label correction. 

\section{Related Works}

In this section, we highlight the related works. There are two key directions we have identified to reduce label cost for time-series classification. 

One deals with semi-supervised learning approach \cite{wei2006semi}, to build accurate classifier using a small set of labelled data, where stopping criteria of training is an open end challenge. Here, authors have neither suggested about label generation of unlabelled data, nor about correction of labels. 
Another type of semi-supervised approaches, known as self-labelling methods, uses a small amount of original labelled set, to classify unlabelled data using the most confident predictions \cite{gonzalez2018}. Self-Training \cite{yarowsky1995} and  Tri-Training \cite{zhou2005} are two such self-labelling techniques, which uses a base classifier, to extend the small labelled set with the most confident instances extracted and classified from the unlabelled set. 

The second, prerequisites heuristic rules which demand expert’s knowledge, and empirical observations to define labels. Like \cite{khattar2019}, devised an algorithm for labelling time-series generated from wearables, using a weak supervision framework and a small percentage of labelled instances. The mechanism incorporates heuristic rules to perform labelling. Finally, a discriminative LSTM model is trained using the weak-supervised labels to obtain performance. Formulation of the heuristic rules for the labelling functions requires domain expertise, which may not be readily available in real-world scenarios.  

We observe, there is recent work that depicts label generation system (Snorkel \cite{ratner2017}) for unstructured data like text. This provides an interface to write labelling functions, exploiting arbitrary heuristics, associated with weak supervision and external knowledge bases. 

Proposed label generation method does not demand any manual intervention. It does not need any heuristics associated with domain knowledge. Our method requires a small set of labelled data. It is inherently adaptive. It exploits learned representation, with a suitable choice of distance measure and performs self-correction to improve the quality of labels. We validate our method across diverse multi-variate, univariate variable length time-series captured from sensors, related to various types of applications like smart health, smart city, manufacturing, and predictive maintenance.


\begin{figure*}[htbp]
\centering
  \includegraphics[height= 2.1 in]{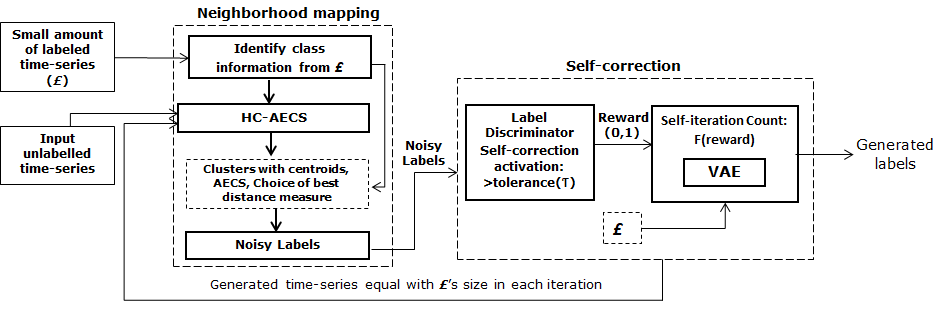}
  \caption{Schematic diagram for proposed approach}
  \label{fig:arch2}
\end{figure*}

\section{Methodology}

\subsection{Selection of representative labelled instances}

The algorithm commences with a complete unlabelled time-series data ($X_u$). From this unlabelled data, a small amount of representative time-series is randomly selected and is annotated by an expert. The number of representative instances $\pounds$, considered for experimentation in our method, is 5\% - 15\% of the complete training dataset. Although, this may vary according to the requirement of the user. Alternatively, a small amount of available labelled representative time-series may be directly provided by the user.

\subsection{Neighborhood Mapping}

Initially, the unlabelled time-series $X_u$ is merged (optional), with the representative time-series $\pounds$, and hierarchical clustering using Auto Encoded Compact Sequence (HC-AECS) is performed on it. The number of clusters to be formed, is kept same as the number of classes, derived from $\pounds$.

\subsubsection{Hierarchical clustering with Auto-Encoded Compact Sequence (HC-AECS)} 

In this method, a compact representation (AECS) of the time-series $X_u$ is learned using a Seq2Seq LSTM multi-layer under-complete auto-encoder \cite{hochreiter1997long,sutskever2014}. An under-complete auto-encoder consists of hidden layers with lower number of nodes than the input/output layer. Hence, the latent representation AECS, has a length much less than the original length of the time-series and learns only the important features from the time-series. After the AECS is learned, Hierarchical clustering \cite{friedman2001} is performed on this compact representation using a method, to find the best choice of distance measure. 
The best distance measure is chosen among Chebyshev(CH), Manhattan(MN) and Mahalanobis(ML) distance using an internal clustering measure Modified Hubert Statistic ($\mathcal{T}$). It evaluates the sum of distance between each pair of time-series, weighted by distance between the cluster centers, which they belong to. The clustering formed by the 3 distance measures is ranked based on $\mathcal{T}$, where the best clustering has the highest value of $\mathcal{T}$. The distance measure corresponding to the best clustering is termed best distance measure. The distance between each pair of time-series and separation between clusters are measured using Mahalanobis distance(ML) to evaluate $\mathcal{T}$.  \\\begin{equation}
\mathcal{T} = \frac{2}{n(n-1)} \sum\limits_{X_i \in X} \sum\limits_{X_j \in X} d(X_i,X_j) d(c_i,c_j),
\end{equation}

\begin{equation}
d(X_i,X_j)= d_{ML}(X_i,X_j) ; d(c_i,c_j) = d_{ML}(c_i,c_j),
\end{equation}

where $C_i$ represents the $i^{th}$ cluster and $c_i$ is the center of cluster $C_{i}$, $d_{ML}(X_i,X_j)$ is the Mahalanobis distance between time-series $X_i$ and $X_j$ and $d_{ML}(c_i,c_j)$ is the Mahalanobis distance between the centres of the clusters to which two time-series $X_i$ and $X_j$ belongs.

The compressed length of the representation reduces the high computation time of hierarchical clustering, which is considered to be its main disadvantage. This mechanism is applicable to both univariate and multi-variate time-series, and, also on variable-length time-series.

\subsubsection{Cluster-class association}

Based on $\pounds$, we aim to associate a class label with each of the clusters formed, using their centroids. It exploits the latent representation AECS of the representative instances to compute their closeness to each cluster centroid. The labels thus formed for instances in $X_u$, are termed as "Noisy labels". 


A distance matrix $dist \in R^{k \times m}$ is computed using the cluster centroids and representative time series $\pounds$ exploiting the best distance measure $d_{best}$.

\begin{equation}
dist[i,j] = d_{best}(cen_i , \pounds_j),
\end{equation}

where $cen_i$ is the centroid of $i^{th}$ cluster and $\pounds_j$ is the AECS of $j^{th}$ representative time-series $\pounds$.

Using $dist$, we find the closest sub-group of $X_u$ for each representative instance. We associate each instance in $\pounds$, to the cluster, whose centroid its learned AECS is nearest to. We define a list $rep\_clus$ which saves each of these associations for $\pounds$. It is defined as :

\begin{equation}
rep\_clus_i = \operatorname*{argmin}_k dist_i ,  \forall i \in m,
\end{equation}

where $k$ denotes the cluster for which $dist[k,i]$ is minimum for $i^{th}$ representative instance and $m$ is the total number of representative instances.

Next for each cluster $j$, we extract the representative instances from $\pounds$ nearest to $cen_j$ and their corresponding class labels in $y_{ins}$. 

The class which occurs most frequently in $y_{ins}$ is associated with cluster $j$ i.e 
\begin{equation}
class_j = Mode(y_{ins}),
\end{equation}

Finally, each unlabelled instance in cluster $j$ are labelled as $class_j$. Detailed algorithm for cluster-class association is described in Algorithm 1. 

\begin{algorithm}
\caption{Cluster-class Association (CCA): Neighborhood Mapping} 
\textbf{Input:} $X_u$: unlabelled time-series, $X_u \in \mathbb{R}^{n \times t \times d}$; \{$\pounds$,$y_r$\}: representative time-series, $\pounds \in \mathbb{R}^{m \times t \times d}$; where $m < n$\\
\textbf{Output:} $y_u$: Associated class labels for unlabelled time-series $X_u$ \\ 
\textbf{begin}
\begin{algorithmic}[1]
\State $x_{AECS} \rightarrow AECS(\pounds)$
\State $k \rightarrow$ No. of unique elements in $y_r$ (Num of classes)
\State $X \rightarrow X_u \cup \pounds$
\State $X_{AECS} ,Clus ,d_{best},Cen \rightarrow HC\_AECS(X)$
\State Compute distance of AECS of each representative time-series to each cluster centroid using best distance measure $d_{best}$ in matrix $dist \in R^{k \times m}$\\
$dist \rightarrow d_{best}(x_{AECS},Cen)$
\State Find the cluster whose centroid has minimum distance for each representative time-series using $dist$ matrix.
$rep\_clus_i \rightarrow \operatorname*{argmin}_k dist_i ,  \forall i \in m$ 
\For{$j = 1,..,k$}
\State Find the representative instances nearest to cluster $j$ 
\State $ins \rightarrow \{i \; \vert \; rep\_clus_i = j\}$ 
\State Find the class labels of instances $ins$ in $y_{ins}$
\State $y_{ins} \rightarrow y_l[ins]$
\State Class having maximum representative instances nearest to centroid of cluster $j$ is associated to cluster $j$.
\State $class_j \rightarrow $ Mode$(y_{ins})$
\EndFor
\State Declare $y_u$ to store the noisy labels for the unlabelled time-series.
\State $y_u \rightarrow \{\}$ 
\For{$j = 1,..,k$}
\State Find the instances in cluster $j$
\State $ins \rightarrow \{i \; \vert \; Clus = j\}$ 
\State Label all the instances in cluster $j$ with $class_j$ for cluster-class association
\State $y_u[ins] \rightarrow class_j$
\EndFor
\State \textbf{return} $y_u$
\end{algorithmic}
\textbf{end}
\end{algorithm}

\subsection{Self-correction module}
The main functionality of self-correction module is to improve purity of the labels generated with each successive iteration. This module comprises following functional components:

\subsubsection{Label Discriminator}
This functional module assesses the quality of labels generated in an iteration. It matches labels generated in two successive iterations. It computes the amount of mismatches of the generated labels in the current iteration with that of the previous, by maintaining an upper bound $\tau$. It triggers reward 0 or 1 to reinforce a new set of representative sample generation. Variations in generated labels in two successive iterations greater than $\tau$, indicates improvement in label quality have not saturated, and, self-correction is to be performed by generating new set of synthetic time-series and augmenting with the previous set of generated $\pounds$, for further refining the purity of labels. In this case, the label discriminator model returns a reward value of 1. If however, the variation of labels generated in the current and previous iteration is within $\tau$ i.e. labels are almost saturated, then the reward value is 0. The detailed algorithm is shown in Algorithm 2.
We use a variable Self Iteration Count($self\_itr\_count$), which is a function of reward(R) and controls the iterations of the self-correction module. 

\begin{algorithm}
\caption{Label Discriminator} 
\textbf{Input:} $NL$: Noisy labels generated in current iteration, $old\_NL$: Noisy labels generated in previous iteration \\
\textbf{Output:} $reward$\\ 
\textbf{begin}
\begin{algorithmic}[1]
\State Checking if noisy labels are identical in consecutive iterations
\If {$Mismatch(NL , old\_NL) \leq \tau$}
\State $reward \rightarrow 0$
\Else
\State $reward \rightarrow 1$
\EndIf
\State \textbf{return} $reward$
\end{algorithmic}
\textbf{end}
\end{algorithm}



\subsubsection{Reinforcing representative sample space using VAE}
Using Variational Auto-encoder(VAE) \cite{doersch2016,kingma2013auto,rezende2014}, we synthetically generate equal amount of representative time-series using the initial small amount of labelled time-series ($\pounds$) in multiple iterations. During the course of proposed multi-stage label generation, amount of representative time-series synthetically generated is varied in every iteration. For example, if $m$ be number of instances in $\pounds$, number of instances generated synthetically in 3 successive iterations are $m$, $2m$, $3m$,.., $nm$ etc. 

VAE is based on generative modelling. It uses Evidence Negative Lower Bound(ELBO) as cost function and a re-parameterization trick (considering $\mu(\pounds)$ and $\Sigma(\pounds)$) to achieve the learning using back-propagation. The loss function for variational auto-encoder is provided in equation (6) where the first term indicates the reconstruction loss and the second term indicates the KL-divergence between the latent distribution and a standard gaussian distribution. Here $\pounds$ is considered as input to VAE and $z$ is the learned latent distribution. \\
\begin{equation}
\log p_{\theta}(\pounds){\geq}E_{q_{\theta}(z|\pounds)}[\log p_{\theta}(\pounds|z)]-D_{KL}(q_{\theta}(z|\pounds)||p(z))
\end{equation}

\begin{figure}[htp]
\centering
  \includegraphics[width=\columnwidth]{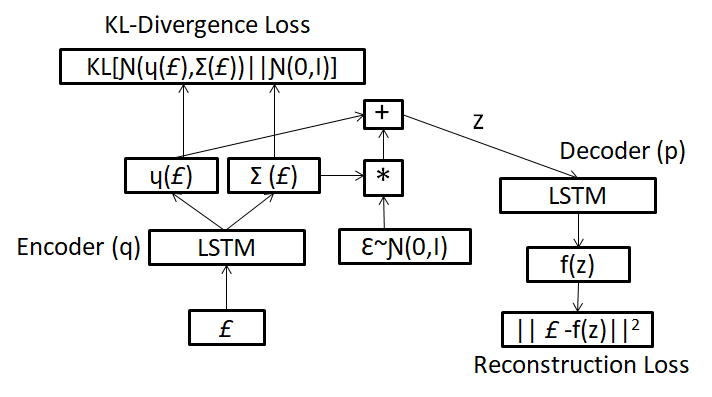}
  \caption{Reinforcing Representative samples($\pounds$) using VAE with enabled back-propagation}
  \label{fig:vae}
\end{figure}

Using first sampling $\epsilon \sim \mathcal{N}(0,I)$, the VAE can sample from a normal distribution $\mathcal{N}(\mu(\pounds),\Sigma(\pounds))$. The latent distribution $z$ can be formalized as: \\
\begin{equation}
z=\mu(\pounds)+\Sigma^{1/2}(\pounds)*\epsilon
\end{equation}
The structure of VAE with the loss terms is shown in figure \ref{fig:vae}. Formalization of the self-correction block is shown in Algorithm 3.
\begin{algorithm}
\caption{Self Correction} 
\textbf{Input:} unlabelled time-series: $X_u$ , $X_u \in \mathbb{R}^{n \times t \times d}$; representative time-series : \{$\pounds$,$y_r$\} , $\pounds \in \mathbb{R}^{m \times t \times d}$ ; \\
\textbf{Output:} $y_{gen} : $ Final generated labels \\ 
\textbf{begin}
\begin{algorithmic}[1]
\State $self\_itr\_count \rightarrow 0$
\State $itr \rightarrow 1$
\State $old\_NL \rightarrow \{\}$
\State $Max\_self\_itr\_count \rightarrow 1$
\While {$self\_itr\_cnt < Max\_self\_itr\_count$}
\If {$itr == 1$ :}
\State $NL \rightarrow CCA(X_u,\pounds,y_r)$
\Else
\State Using a VAE we generate new representative samples exploiting given small amount of labelled samples from each class.
\State $x_{gen},y_{gen} \rightarrow VAE(\pounds)$ 
\State We add the generated instances in set of representative instances
\State $new\_\pounds \rightarrow \pounds \cup x_{gen} , newy_r \rightarrow y_r \cup y_{gen}$
\State $NL \rightarrow CCA(X_u,new\_\pounds,newy_r)$
\EndIf 
\State Checking quality of labels and reward generation
\State $reward \rightarrow$ Label\_discriminator($NL,old\_NL$)
\State $Max\_self\_itr\_count \rightarrow reward+reward$
\If {$reward == 0$}
\State $self\_itr\_cnt \rightarrow self\_itr\_cnt + 1$ 
\Else
\State  $self\_itr\_cnt \rightarrow 0$
\EndIf
\State $old\_NL \rightarrow NL$
\State $itr \rightarrow itr + 1$
\EndWhile
\State $y_{gen} \rightarrow NL$
\State \textbf{return} $y_{gen}$
\end{algorithmic}
\textbf{end}
\end{algorithm}

\section{Validation}

\begin{figure}[htbp!]
\centering
  \includegraphics[width=\columnwidth]{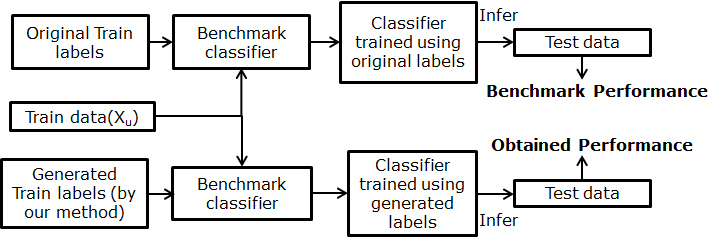}
  \caption{Validation Framework}
  \label{fig:val_arc}
\end{figure}

In this section we present the validation approach. We train different benchmark classifier models like BOSS \cite{schafer2015boss}, Random Forest \cite{breiman2001}, MLSTM-FCN \cite{karim2019} and MLP \cite{friedman2001} using train data $X_u$ and generated labels for $X_u$ using proposed method. Subsequently, we infer the hidden test data on the trained model to obtain the final performance. We compare this performance with benchmark accuracy using same benchmark classifiers. The benchmark accuracy is obtained by applying same hidden test data, using the model trained with $X_u$ and original training labels. The validation framework for proposed approach is depicted in Figure \ref{fig:val_arc}.

\section{Experimental Analysis}

In this section, we present experimental results demonstrating capability of proposed automated label generation using learned representation along with self-correction and choice of best distance measure. 

\subsection{Dataset Description}

In this work univariate, multi-variate, variable length time-series across diverse IoT application domains generated by sensors like ECG, accelerometers, process control sensors etc. from UCR \cite{dau2019ucr} and UCI \cite{asuncion2007uci} archive have been considered. A small percentage (5\%-15\%) of representative time-series is constructed by randomly chosen data from the train set. This can alternatively be obtained by expert validation separately. The test set is hidden and reserved, on which the method is evaluated upon.

\begin{table}[htp]
\centering
\caption{Dataset Description table}
\label{uni_dd}
\resizebox{\columnwidth}{!}{\begin{tabular}{cccccc}
\hline
\textbf{Sensor} & \textbf{Dataset} & \textbf{\#Train} & \textbf{\#Test} & \textbf{Length} & \textbf{\#class} \\ \hline
Accelerometer & SonyAIBORobotSurface2 & 27	& 953 &	65 & 2 \\ 
              & Earthquakes & 322 & 139 & 512 & 2 \\ \hline
Motion & ToeSegmentation1 & 40 &	228 & 277 &	2 \\ 
Capture & ToeSegmentation2 &	36 & 130 &	343 & 2 \\ 
Camera & GunPoint & 50 &	150 & 150 & 2 \\ \hline
Process & MoteStrain & 20 & 1252 & 84 & 2 \\ 
Sensors & Wafer &  298 &	896 & 104-198 & 2 \\ \hline
Camera & DistalPhalanxOAG & 400 & 139 & 80 & 3 \\ 
	   & ProximalPhalanxOAG & 400 & 205 & 80 & 3 \\ \hline
ECG & TwoLeadECG & 23 & 1139 & 82 & 2 \\ 
    & ECG5000 & 500 & 4500 & 140 & 5 \\ \hline
Elec. Meter & PowerCons & 180 & 180 & 144 & 2 \\ \hline

\end{tabular}}
\end{table}

\begin{figure*}[htbp]
\centering
\begin{subfigure}{.3\textwidth}
\centering
  \includegraphics[width=55mm,height=40mm]{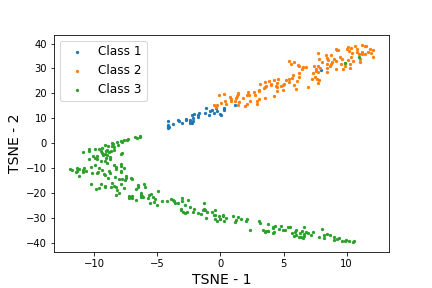}
  \caption{Iteration 1}
  \label{fig:sub1}
\end{subfigure}%
\begin{subfigure}{.3\textwidth}  
\centering
  \includegraphics[width=55mm,height=40mm]{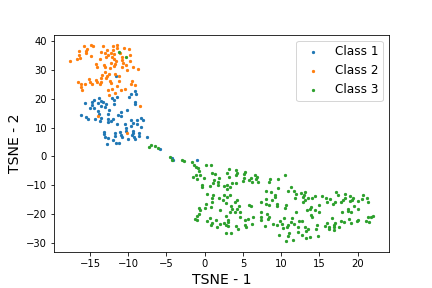}
  \caption{Iteration 2}
  \label{fig:sub2}
\end{subfigure}
\begin{subfigure}{.3\textwidth}  
\centering
  \includegraphics[width=55mm,height=40mm]{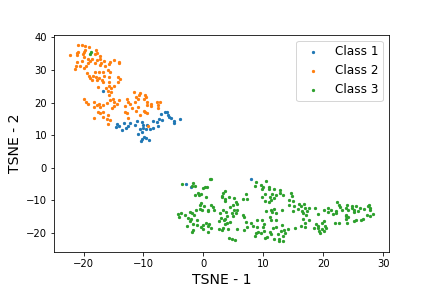}
  \caption{Iteration 3}
  \label{fig:sub3}
\end{subfigure}
\caption{TSNE-plot for AECS depicting self-correction of generated labels in successive iterations for DistalPhalanxOAG}
\label{fig:dpoag}
\end{figure*}

\begin{figure*}[htbp]
\centering
\begin{subfigure}{.3\textwidth}
\centering
  \includegraphics[width=55mm,height=40mm]{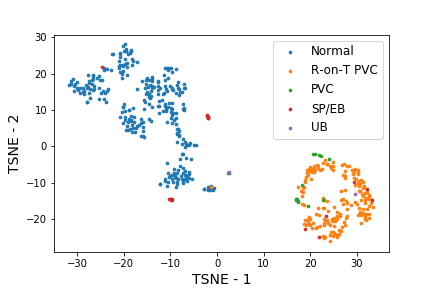}
  \caption{Iteration 1}
  \label{fig:sub1}
\end{subfigure}%
\begin{subfigure}{.3\textwidth}  
\centering
  \includegraphics[width=55mm,height=40mm]{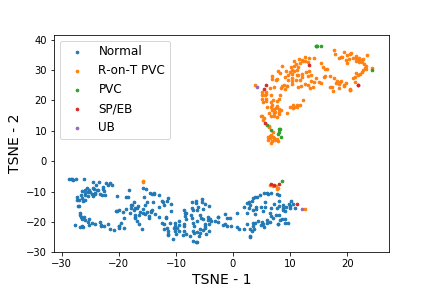}
  \caption{Iteration 2}
  \label{fig:sub2}
\end{subfigure}
\begin{subfigure}{.3\textwidth}  
\centering
  \includegraphics[width=55mm,height=40mm]{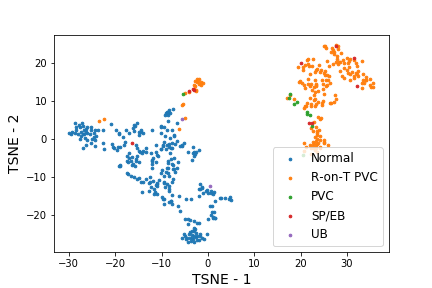}
  \caption{Iteration 3}
  \label{fig:sub3}
\end{subfigure}
\caption{TSNE-plot for AECS depicting self-correction of generated labels in successive iterations for ECG5000 dataset}
\label{fig:ecg}
\end{figure*}

\subsection{Neighborhood mapping}
\label{subsec:cca}

For neighborhood mapping, we use HC-AECS as the clustering method, as it recommends a best distance measure and performs robustly across a wide variety of time-series. Furthermore, it requires much less computation time, as clustering is applied on the compressed representation instead of the raw time-series \cite{bandyopadhyay2021}. It needs approximately 27 times less computation time, as compared to benchmark time-series clustering approach HC-DDTW \cite{luczak2016}, which performs clustering on raw time-series.  

We consider DistalPhalanxOAG, as an exemplary univariate time-series, to demonstrate our approach. This data consists of bone-outlines, in form of univariate time-series, for classification of subject's age among 3 classes/age groups(0-6, 7-12 and 13-19 years). It has 400 train instances of length 80 each ($X_u \in \mathcal{R}^{400 \times 80})$. We randomly select a small percentage of total training instances as representative labelled data $\pounds$ considering equal number of instances from each class. We generate auto encoded compact sequence on $X_u$, which produces a compressed representation of length 12 for each instance $X_{AECS} \in \mathcal{R}^{400 \times 12}$. Applying Hierarchical clustering on $X_{AECS}$, we observe Chebyshev distance produces the highest Modified Hubert Statistic($\mathcal{T}$) among the 3 distance measures ($\mathcal{T}_{CH}=\textbf{0.574}, \mathcal{T}_{MN}=0.571, \mathcal{T}_{ML}=0.572$). Hence, Chebyshev distance is chosen for forming the clusters and performing the cluster-class association. We vary the percentage of $\pounds$ as 5\%, 10\% and 15\%, and generate 3 separate sets of training labels. 


We also consider a univariate time-series ECG5000 for demonstration. It is a dataset for classification of normal and different types of abnormal ECG signals. It consists of 500 train instances where each time series length is 140 ($X_u \in \mathcal{R}^{500 \times 140}$). Applying HC-AECS, we obtain a compressed representation $X_{AECS} \in \mathcal{R}^{500 \times 12}$ for the time-series.  We observe Manhattan distance produces the highest Modified Hubert Statistic($\mathcal{T}$) among the 3 distance measures ($\mathcal{T}_{CH}=0.301, \mathcal{T}_{MN}=\textbf{0.302}, \mathcal{T}_{ML}=0.301$) and hence is considered for cluster-class association.

\begin{table*}[htb!]
\centering
\caption{Accuracy comparison of proposed method (with varying labelled data \% considered) with generated labels vs. original labels using Benchmark classifier} 
\label{comp}
\begin{tabular}{ccccccc}
\hline

\textbf{Sensor} & \textbf{Dataset} & \textbf{Benchmark}    & \textbf{Benchmark} & \multicolumn{3}{c}{\textbf{Proposed Method}}  \\ \cline{4-7}
& & \textbf{Classifier}    & \textbf{100\%} &   \textbf{5\%}    &    \textbf{10\%}   &   \textbf{15\%} \\ \hline      

Accelerometer & SonyAIBORobotSurface2 & BOSS  & 0.831 & 0.738 & 0.744 & 0.786 \\ 
              & Earthquakes  & RandF  & 0.748 & \textbf{0.748} & \textbf{0.755} & \textbf{0.755} \\ \hline 
Motion        &  ToeSegmentation1      & BOSS  & 0.851 & 0.728 & \textbf{0.868} & \textbf{0.868} \\ 
Capture       & ToeSegmentation2      & BOSS  & 0.923  & 0.762 & 0.885 & \textbf{0.923}\\ 
Camera        & GunPoint			  & BOSS & 0.980 & 0.700 & 0.787 & 0.913  \\ \hline
Process       & Wafer UCI	& MLSTM-FCN & 0.909 & \textbf{0.965} & \textbf{0.965} & \textbf{0.965} \\ 
Sensors       & MoteStrain			  & BOSS & 0.895 & 0.602 & 0.809 & 0.841   \\ \hline
ECG          & TwoLeadECG			  & BOSS & 0.895 & 0.500 & 0.672 & 0.870  \\ 
			 & ECG5000				& MLP  & 0.940 & 0.927  & 0.935 &  0.923    \\ \hline
Camera       & DistalPhalanxOAG	  & MLP  & 0.727 & 0.612 & 0.712 & 0.719 \\ 
          & ProximalPhalanxOAG    & MLP   & 0.844 & \textbf{0.854} & \textbf{0.854} & \textbf{0.854} \\ \hline
Elec. Meter & PowerCons             & BOSS  & 0.866 & 0.739 & 0.800 & 0.850 \\ \hline
\end{tabular}
\end{table*}

\subsection{Analysis on Self-correction module}

We use a single layer LSTM encoder-decoder model for implementing the VAE. The two parameters in the latent space - mean($\mu$) and standard deviation($\Sigma$) are implemented as Dense layers. A normal latent distribution with $\mu=0$ and $\sigma=1$ is learnt using the VAE where the length of each latent sample is same as the length of the original time-series. RMSProp optimiser \cite{hinton2012} with a learning rate of 0.003 have been used for training the model. 

Labels generated in successive iterations are considered to be saturated when the percentage of mismatch among them is less than or equal to 5\%. Here we consider the value of threshold $\tau$ as 0.05 in the label discriminator module.

Continuing our experimentation using exemplary dataset DistalPhalanxOAG in section \ref{subsec:cca}, we illustrate how the quality of labels improves with each successive iteration using self-correction. Figure \ref{fig:dpoag} shows the iteration-wise improvement for DistalPhalanxOAG using our proposed method exploiting 15\% of labelled data. Similarly for ECG5000 dataset, the improvement of the label quality in each successive iteration for ECG5000 using 10\% labelled data is shown in figure \ref{fig:ecg}.

\subsection{Results}

In this section, we present the results of our proposed method. We compare in Table \ref{comp}, performance of benchmark classifiers using original training labels and generated labels using our proposed model. 

We have exploited 5\% to 15\% of training data as $\pounds$ for label generation of entire training set for experimentation. Here, we observe in five out of the twelve datasets, the performance of the classifiers using the generated labels is higher or equal to the performance using original training labels (marked bold in Table \ref{comp}). We see, among three of the five datasets, viz. Earthquakes, Wafer and ProximalPhalanxOAG, obtained performance using just 5\% of labelled data, is higher/equal to benchmark performance. Furthermore, in four other datasets, the difference between the performance of the classifier using original training labels and generated labels, is within an acceptable threshold (3\%). Using maximum of 15\% labelled data, the accuracy of our proposed approach, on average, differs by just 1.08\%, from the performance using original training labels.

\subsubsection{Comparison with State-of-the-art method}
We compare proposed method with a State-of-the-art self-labelling technique, Tri-Training (TrT) \cite{zhou2005}, on ten univariate time-series from UCR archive. For comparison, 15\% labelled data from the training set have been used for self-labelling for both the methods and the performance have been evaluated on the same test set. KNN and Decision Tree have been used as the base classifiers for both the methods. Table \ref{comp_trit} depicts the detailed comparison of our approach with Tri-Training. We observe, using both KNN and Decision Tree as base classifiers, proposed method outperforms Tri-Training in seven out of ten datasets i.e. 70\% of cases. Furthermore, we perform Student's t-test \cite{owen1965} to determine, if there is any significant statistical difference against evaluated performance of proposed method. We observe in five of the datasets, the improvement is statistically significant with 95\% confidence level \cite{johnson2000}, highlighting the global accuracy of the proposed method is superior than TrT.

\begin{table}[htb!]
\centering
\caption{Comparison of proposed method with Tri-Training with KNN and Decision Tree(DT) as base classifiers (based on accuracy)} 
\label{comp_trit}
\resizebox{\columnwidth}{!}{\begin{tabular}{c|cc|cc}
\hline
\textbf{Dataset} & \multicolumn{2}{c|}{\textbf{Classifier: KNN}} & \multicolumn{2}{c}{\textbf{Classifier: DT}} \\ 
 &	\textbf{TrT} &	\textbf{Proposed} & \textbf{TrT} &	\textbf{Proposed} \\ \hline
SonyAIBORobotSurface2 &	0.605 &	\textbf{0.754} &	0.555 &	\textbf{0.606} \\
TwoLeadECG            & 0.539 &	\textbf{0.687} &	0.662 &	\textbf{0.715} \\
Earthquakes & 	\textbf{0.705} &	0.631 &	\textbf{0.698} &	0.669 \\
MoteStrain &	\textbf{0.7} &	0.581 &	\textbf{0.593} &	0.562 \\
PowerCons &	0.683 &	\textbf{0.694} &	\textbf{0.872} &	0.667 \\
ECG5000 &	0.901 &	\textbf{0.913} &	0.871 &	\textbf{0.883} \\
ToeSegmentation1 &	0.465 &	\textbf{0.535} &	0.478 &	\textbf{0.526} \\
ToeSegmentation2 &	\textbf{0.723} &	0.623 &	0.431 &	\textbf{0.531} \\
ProximalPhalanxOAG &	0.756 &	\textbf{0.854} &	0.678 &	\textbf{0.854} \\
DistalPhalanxOAG &	0.633 &	\textbf{0.654} &	0.611 &	\textbf{0.633}\\ \hline \hline
Wins            & 3/10  & \textbf{7/10} & 3/10 & \textbf{7/10} \\ \hline \hline

\end{tabular}}
\end{table}

\section{Conclusion}

In this paper, we have presented an automated label generation method for time-series classification with representation learning using a small amount of labelled data. Our multi-stage label generation method self-corrects in iterations by using VAE (Variational auto encoder) based generative modelling to improve the quality of labels by learning latent structure of representative labelled data ($\pounds$) and increase equal amount of $\pounds$ guided by a label discriminator module. Proposed method uses a robust hierarchical clustering approach using a combination of hierarchical clustering with multi-layer under complete Seq2Seq auto-encoder representation, AECS (Auto-Encoded Compact Sequence) of both univariate and multivariate time-series along with the choice of best distance measure CH, MA, and ML. The selected best distance measure is used for performing Neighborhood mapping w.r.t learned compact representation of $\pounds$. 

We have performed extensive analysis considering real-world univariate, multivariate, and variable length time-series related to different application domains like smart health, smart city, manufacturing  from UCR and UCI time series classification archives.  Experimental results show that generated labels of unlabelled time-series indeed perform very closely w.r.t the original labelled time-series, and in some cases, it outperforms proving to achieve label correction. We observe, in 42\% (five out of twelve) of datasets, the proposed method using 5 to 15\% labelled data, outperforms the benchmark performance. On average,  using maximum of 15\% labelled data, the difference between performance of proposed method with the benchmark performance is very low (1.08\%). We have compared our approach with state-of-the-art techniques. We observe our approach outperforms in seven out of ten datasets (70\% of cases) w.r.t Tri-Training, a prior work using the self-labelling method. 

The proposed method does not need any manual intervention or any kind of heuristic associated with domain knowledge to generate labels. Only usage of a small amount of representative labelled data reduces the cost of expert knowledge to a large extent.

\bibliographystyle{named}
\bibliography{ijcai21}

\begin{thebibliography}{}

\bibitem[\protect\citeauthoryear{Asuncion and Newman}{2007}]{asuncion2007uci}
Arthur Asuncion and David Newman.
\newblock Uci machine learning repository, 2007.

\bibitem[\protect\citeauthoryear{Bandyopadhyay \bgroup \em et al.\egroup
  }{2021}]{bandyopadhyay2021}
Soma Bandyopadhyay, Anish Datta, and Arpan Pal.
\newblock Hierarchical clustering using auto-encoded compact representation for
  time-series analysis.
\newblock {\em arXiv preprint arXiv:2101.03742}, 2021.

\bibitem[\protect\citeauthoryear{Breiman}{2001}]{breiman2001}
Leo Breiman.
\newblock Random forests.
\newblock {\em Machine learning}, 45(1):5--32, 2001.

\bibitem[\protect\citeauthoryear{Dau \bgroup \em et al.\egroup
  }{2019}]{dau2019ucr}
Hoang~Anh Dau, Anthony Bagnall, Kaveh Kamgar, Chin-Chia~Michael Yeh, Yan Zhu,
  Shaghayegh Gharghabi, Chotirat~Ann Ratanamahatana, and Eamonn Keogh.
\newblock The ucr time series archive.
\newblock {\em IEEE/CAA Journal of Automatica Sinica}, 6(6):1293--1305, 2019.

\bibitem[\protect\citeauthoryear{Doersch}{2016}]{doersch2016}
Carl Doersch.
\newblock Tutorial on variational autoencoders.
\newblock {\em arXiv preprint arXiv:1606.05908}, 2016.

\bibitem[\protect\citeauthoryear{Friedman \bgroup \em et al.\egroup
  }{2001}]{friedman2001}
Jerome Friedman, Trevor Hastie, Robert Tibshirani, et~al.
\newblock {\em The elements of statistical learning}, volume~1.
\newblock Springer series in statistics New York, 2001.

\bibitem[\protect\citeauthoryear{Gonz{\'a}lez \bgroup \em et al.\egroup
  }{2018}]{gonzalez2018}
Mabel Gonz{\'a}lez, Christoph Bergmeir, Isaac Triguero, Yanet Rodr{\'\i}guez,
  and Jos{\'e}~M Ben{\'\i}tez.
\newblock Self-labeling techniques for semi-supervised time series
  classification: an empirical study.
\newblock {\em Knowledge and Information Systems}, 55(2):493--528, 2018.

\bibitem[\protect\citeauthoryear{Hinton \bgroup \em et al.\egroup
  }{2012}]{hinton2012}
Geoffrey Hinton, Nitish Srivastava, and Kevin Swersky.
\newblock Neural networks for machine learning lecture 6a overview of
  mini-batch gradient descent.
\newblock {\em Cited on}, 14(8), 2012.

\bibitem[\protect\citeauthoryear{Hochreiter and
  Schmidhuber}{1997}]{hochreiter1997long}
Sepp Hochreiter and J{\"u}rgen Schmidhuber.
\newblock Long short-term memory.
\newblock {\em Neural computation}, 9(8):1735--1780, 1997.

\bibitem[\protect\citeauthoryear{Hubert and Arabie}{1985}]{hubert1985}
Lawrence Hubert and Phipps Arabie.
\newblock Comparing partitions.
\newblock {\em Journal of classification}, 2(1):193--218, 1985.

\bibitem[\protect\citeauthoryear{Johnson \bgroup \em et al.\egroup
  }{2000}]{johnson2000}
Richard~A Johnson, Irwin Miller, and John~E Freund.
\newblock {\em Probability and statistics for engineers}, volume 2000.
\newblock Pearson Education London, 2000.

\bibitem[\protect\citeauthoryear{Karim \bgroup \em et al.\egroup
  }{2019}]{karim2019}
Fazle Karim, Somshubra Majumdar, Houshang Darabi, and Samuel Harford.
\newblock Multivariate lstm-fcns for time series classification.
\newblock {\em Neural Networks}, 116:237--245, 2019.

\bibitem[\protect\citeauthoryear{Khattar \bgroup \em et al.\egroup
  }{2019}]{khattar2019}
Saelig Khattar, Hannah O’Day, Paroma Varma, Jason Fries, Jennifer Hicks,
  Scott Delp, Helen Bronte-Stewart, and Chris Re.
\newblock Multi-frame weak supervision to label wearable sensor data.
\newblock In {\em ICML Time Series Workshop}, 2019.

\bibitem[\protect\citeauthoryear{Kingma and Welling}{2013}]{kingma2013auto}
Diederik~P Kingma and Max Welling.
\newblock Auto-encoding variational bayes.
\newblock {\em arXiv preprint arXiv:1312.6114}, 2013.

\bibitem[\protect\citeauthoryear{{\L}uczak}{2016}]{luczak2016}
Maciej {\L}uczak.
\newblock Hierarchical clustering of time series data with parametric
  derivative dynamic time warping.
\newblock {\em Expert Systems with Applications}, 62:116--130, 2016.

\bibitem[\protect\citeauthoryear{Owen}{1965}]{owen1965}
DB~Owen.
\newblock The power of student's t-test.
\newblock {\em Journal of the American Statistical Association},
  60(309):320--333, 1965.

\bibitem[\protect\citeauthoryear{Ratner \bgroup \em et al.\egroup
  }{2017}]{ratner2017}
Alexander Ratner, Stephen~H Bach, Henry Ehrenberg, Jason Fries, Sen Wu, and
  Christopher R{\'e}.
\newblock Snorkel: Rapid training data creation with weak supervision.
\newblock In {\em Proceedings of the VLDB Endowment. International Conference
  on Very Large Data Bases}, volume~11, page 269. NIH Public Access, 2017.

\bibitem[\protect\citeauthoryear{Rezende \bgroup \em et al.\egroup
  }{2014}]{rezende2014}
Danilo~Jimenez Rezende, Shakir Mohamed, and Daan Wierstra.
\newblock Stochastic backpropagation and approximate inference in deep
  generative models.
\newblock In {\em International conference on machine learning}, pages
  1278--1286. PMLR, 2014.

\bibitem[\protect\citeauthoryear{Sch{\"a}fer}{2015}]{schafer2015boss}
Patrick Sch{\"a}fer.
\newblock The boss is concerned with time series classification in the presence
  of noise.
\newblock {\em Data Mining and Knowledge Discovery}, 29(6):1505--1530, 2015.

\bibitem[\protect\citeauthoryear{Sutskever \bgroup \em et al.\egroup
  }{2014}]{sutskever2014}
Ilya Sutskever, Oriol Vinyals, and Quoc~V Le.
\newblock Sequence to sequence learning with neural networks.
\newblock {\em arXiv preprint arXiv:1409.3215}, 2014.

\bibitem[\protect\citeauthoryear{Wei and Keogh}{2006}]{wei2006semi}
Li~Wei and Eamonn Keogh.
\newblock Semi-supervised time series classification.
\newblock In {\em Proceedings of the 12th ACM SIGKDD international conference
  on Knowledge discovery and data mining}, pages 748--753, 2006.

\bibitem[\protect\citeauthoryear{Yarowsky}{1995}]{yarowsky1995}
David Yarowsky.
\newblock Unsupervised word sense disambiguation rivaling supervised methods.
\newblock In {\em 33rd annual meeting of the association for computational
  linguistics}, pages 189--196, 1995.

\bibitem[\protect\citeauthoryear{Zhou and Li}{2005}]{zhou2005}
Zhi-Hua Zhou and Ming Li.
\newblock Tri-training: Exploiting unlabeled data using three classifiers.
\newblock {\em IEEE Transactions on knowledge and Data Engineering},
  17(11):1529--1541, 2005.

\end{thebibliography}

\end{document}